\RequirePackage[2020-02-02]{latexrelease}
\documentclass{clv3}

\usepackage{hyperref}
\usepackage{xcolor}
\definecolor{darkblue}{rgb}{0, 0, 0.5}
\hypersetup{colorlinks=true,citecolor=darkblue, linkcolor=darkblue, urlcolor=darkblue}

\begin{document}
\issue{1}{1}{2022}

\dochead{The Role of Explanatory Value in Natural Language Processing}


\runningtitle{Explanatory Value in NLP}

\runningauthor{Kees van Deemter}


\author{Kees van Deemter.}
\affil{Utrecht University}




\maketitle

\begin{abstract}
A key aim of science is explanation, yet the idea of explaining language phenomena has taken a backseat in mainstream Natural Language Processing (NLP) and many other areas of Artificial Intelligence. I argue that explanation of linguistic behaviour should be a main goal of NLP, and that this is not the same as making NLP models ``explainable". To illustrate these ideas, some recent models of human language production are compared with each other. I conclude by asking what it would mean for NLP research and institutional policies if our community took explanatory value seriously, while heeding some possible pitfalls.
\end{abstract}


\section{Introduction}

%

In this short paper, I will argue that much recent work in Natural Language Processing (NLP) has focussed too narrowly on the {\em performance} of its models, as measured by various intrinsic or extrinsic evaluations, neglecting some vitally important other dimensions of model quality, which I will collect loosely under the header "explanatory value".\footnote{Although I am conscious that this use of the term ``explanatory" is broader than its daily usage, I will argue in section 2 that it makes sense to use the term in this way.} Before explaining what I mean, let me set the scene.

Broadly, NLP can be pursued in three mutually overlapping ways, which emphasize different aspects of our work. First, there is NLP as Engineering, where NLP models are built primarily to serve some practical goal, answering a need that exists in society.
Second, there is what might be called NLP-as-Mathematics, which studies algorithms and models in their own right, comparing them and developing new ones. Finally, there is NLP-as-Science, where models are constructed with the aim of expressing, testing, and ultimately enhancing humankind's grasp of human language and language use, because computational models offer a level of explicitness and detail that other theories of language often lack. For example, a sentiment analysis model 
may be seen as a highly explicit theory of the ways in which psychological states manifest themselves in language; a summarisation model 
can be seen as a theory of what is most informative in a text; a Machine Translation model may be seen as a theory of translation. 

I will focus primarily on NLP-as-Science arguing that, if explanation is our aim, then this should include several dimensions. In particular, if we rely solely on the performance of our models, then we risk building models that are {\em ad hoc}, that are unwieldy, that are difficult to link with existing insights, 
and that do not allow us to answer counterfactual questions such as, "How well would this model perform if we applied it to texts of a different genre?"\\[1ex]
%
%
One might argue that a lot of work in NLP focusses on explanation already, because it aims for explainability (e.g., \cite{ghas}).
%
This argument, however, confuses (1) explaining a natural phenomenon (e.g., an aspect of language use) and (2) explaining a model (i.e., a piece of software). Explanation is about "What principles underlie this phenomenon?", whereas explainability is about "Why did this model make these decisions?" The difference is starkest when the model doesn't match the phenomena very well. 
Suppose you had a model for classifying student essays into good (pass) and bad (fail). Suppose the model had terrible performance but excellent explainability.\footnote{For instance, via computer-generated ``rationales" that highlight text fragments that were particularly important for each classification decision, e.g. \citet{lei}.} The model would thus be highly explainable, and this could be useful for a stakeholder who wonders whether to trust its decisions, or a developer wanting to improve it. Yet these explanations would not tell us what makes an essay good or bad because (we assumed) the model does not know the difference. 
%
%
Similar things can be said about the idea that model evaluation should include a systematic analysis of errors (e.g., \cite{mcke}; \cite{Ribe}).
%
The importance of finding out what a model gets right and wrong can hardly be overstated but, by itself, it can only shed light on the model, not the language phenomena we are seeking to understand.
%
%
%



\section{Dimensions of Explanatory Value}

Explanation plays a key role throughout the sciences 
\cite{popp, over, wood} and in daily life \cite{lomb}. 
Accordingly, many disciplines have seen lively discussion of what it means for a theory or model to offer a good explanation of data, but in recent NLP there has been little discussion about such matters. Consequently, it is unclear what makes a good explanation, and whether explanation even matters in NLP. Based on a liberal borrowing from other disciplines, I will discuss what different dimensions explanation can involve, what these might mean for assessing NLP models, and what the implications would be if our community took them seriously.\\[1ex]
%
%
%
%
%
{\bf Performance.} One dimension of explanatory value is {\em performance}, which 
includes metrics such as Precision, Recall, DICE, BLEU, Moverscore \cite{sai}, \cite{celi}, which allow researchers to compare a model's predictions with a gold standard. 
Evaluations based on human judgments 
(e.g., \cite{lee}) or task-based evaluations are varieties of performance likewise.
%
Performance is naturally thought of as a component of explanatory value. For instance, 
if model A has better performance than B, then {\em other things being equal}, A has greater explanatory value than B. If a model does not allow performance to be assessed at all, then I will consider such models to have low performance.\\[2ex]
%
%
A natural complement to performance is a group of factors I will call ``support". Simply put, when a model is tested in an experiment, support is any evidence for the model other than the evidence from that particular experiment. I will distinguish between (what I will call) direct and indirect support.\\[1ex]
{\bf Direct support.} Direct support assesses a model's ability to make accurate predictions for unseen data, and to generalize to related tasks and different contexts. It comes from a plurality of broadly similar experiments.\footnote{This includes repetitions, reproductions, and replications, using the terminology of the ACM at  www.acm.org/publications/policies/artifact-review-and-badging-current .} Suppose a caption generation model\footnote{Caption generation models are Vision and Language models that generate textual captions for visual images \cite{hodo,agra}.} was tested on a set of holiday snaps, showing good performance. Direct support could include an experiment on another collection of holiday snaps, an experiment with a different type of images, or a probe that investigates whether the success of the model is due to some accidental feature of the dataset. Similarly, if a Machine Translation model is tested on a new language pair, then this can be seen as offering direct support to the model. 
Direct support views language corpora as data {\em samples} that are only of interest to the extent that they are representative of a wider {\em population} of data. This implies that we should ask ourselves what type of language use we want our corpora to be representative of, and be on our guard against confounding variables (i.e., accidental circumstances that may have affected our results \citet{ribe2}).\\[2ex]
%
%
{\bf Indirect support.} A dominant theme in the Philosophy of Science is that explanation should involve a reduction of the unknown (e.g., some previously unobserved facts) to something already known, such as an existing law or insight or model \cite{hemp}. 
These ideas lie at the heart of the scientific enterprise (see e.g. \citet{hepb}), understood as ``the attempt to understand the world around us" (\citet{Leve}, speaking about Artificial Intelligence).





For example, suppose a study finds that a medicine $X$ damages patients' liver. Suppose we also know $X$ contains a molecule $Y$ which is known to be toxic, then this existing insight offers indirect support to the finding about side effects. Likewise, if an aspect of language use can be shown to enhance or speed up communication, this can lend additional credence to a model that incorporates this aspect. Support can also be negative. For example, if a physics model explains certain observations by positing ``action at a distance", (i.e., where an object can be affected by another object without being ``touched" by it in any way) then the difficulty of making sense of that idea has been seen as diminishing the value of the model \cite{berk}. 

Indirect support is essential to what is called explanation in everyday parlance, and can even give us a sense that we ``understand" the underlying mechanism (for instance when we know that the above molecule $Y$ kills a particular liver enzyme), but the idea of mechanism has been notoriously difficult to underpin \cite{crav}. Indirect support in NLP can sometimes take the form of cognitive plausibility (although it does not have to, see e.g. \citet{harn}). Suppose, for example, a text comprehension model uses an algorithm that is NP-complete, then this makes it implausible as a model of human behaviour because it suggests a brain mechanism that would be so time consuming that it could not work in practice.
When indirect support $I$ is invoked, two key questions are in order: How certain are we about $I$ (e.g., what is the evidence or reasoning underlying $I$?), and To what extent would $I$ (if correct) support the model?  
%
Most scientists accept that higher principles have a role to play, though an assessments of whether a principle is rightfully invoked can be a matter of vigorous debate. For instance, action-at-a-distance has become an accepted part of physics despite being seen as implausible for a long time.\\[2ex]
%
%
{\bf Parsimony.} Parsimony is the idea that a simpler model is a better model. Parsimony is closely related to Ockham's Razor and to the idea that models should be as {\em elegant} as possible (see e.g. \citet{gree}), which counts philosophers and physicists such a Karl Popper and Paul Dirac among its early proponents. Parsimony is an aspect of explanatory value because if we do not insist on some form of parsimony, a model could be deemed to be highly explanatory even if it was nothing more than a large collection of isolated facts or rules without any attempt at generalisation. 

Regarding the question of how scientific elegance should defined, there are different views. In particular, parsimony can concern different aspects of the model (see e.g. \citet{fitz}, section 3). Accounts of parsimony that focus on the number of postulates employed by the model, for example, have been defended on the grounds that more parsimonious models have a greater probability of being true.
 %
%
A type of parsimony more relevant to NLP says that, even if two models cannot (or: not very clearly) be distinguished in terms of their performance, then if one is simpler than the other, the simpler model should be preferred. This type of parsimony is routinely used, and sometimes defended explicitly, by syntacticians \cite{brod,akma}, among others.\footnote{The opening chapter of \citet{akma} uses this example: {\em X liked you} is assigned the underlying form {\em X did like you}, because this allows one to generate tag questions ({\em X liked you, didn't he?}), negated sentences ({\em X did not like you}) and emphatic sentences ({\em X did like you}) using one and the same mechanism, thereby minimizing the complexity of the grammar.}

Invoking parsimony can be risky, particularly if a complex model is dismissed that has better performance than its competitors. Physicists such as Sabine Hossenfelder believe elegance has played too large a role in discussions of string theory 
\cite{hoss}. Nonetheless, the idea that a lack of parsimony can diminish the value of a theory is widely accepted. We will count it as a fourth dimension of explanatory value.\\[1ex]
A {\bf Bayesian perspective} on the progression of science (e.g. \cite{jayn}) may help to clarify these dimensions. Let $D$ be the data obtained when a model $M$ is tested, and $X$ is everything else we know, including {\em indirect support} for or against $M$. Then {\em performance} of a probabilistic model $M$ can be seen as $P(D | M, $X$)$, the probability $D$ would have if $M$ were true. 
What one is typically interested in is
$P( M | D,X)$, the probability of $M$ given $D$ and $X$.
{\em Direct support} is Bayesian update, where more and more data sets $D_1,..,D_n$ are brought to bear, yielding $P( M | D_1, .., D_n, X)$. Even {\em parsimony} can, at least in principle, be captured along Bayesian lines, by using Solomonoff's Prior \cite{solo, hutt}, which assesses the complexity of a model by measuring its {\em a priori} probability (i.e., the probability of the model before any data are considered).

\section{Case study: Two types of Referring Expressions Generation}
\label{case-study}

To illustrate both the usefulness and the pitfalls of assessing the explanatory value of NLP models, I examine two types of referring expressions generation (REG). I choose REG because referring is an essential part of human communication that has been studied from many different angles, using very different types of models; moreover, the performance of REG models has been tested extensively, and the outcomes of these tests will inform our discussion of the explanatory value of these models.

\subsection{Generating one-shot Referring Expressions} 

{\em One-shot REG} has been much studied in NLP (\cite{dale}, \cite{reit}, \cite{krah}, \cite{yu}, \cite{luo}): the research question is, given a ``scene" composed of objects, and without any linguistic context (hence ``one-shot" REG), what properties do human speakers select when they refer to one of the objects in the scene? The patterns observed here are far from trivial, and sometimes counter-intuitive \cite{kees}. Here I concentrate on a class of models that emerged from controlled experiments involving simple artificial scenes whose objects have well-understood properties (shape, colour, size, etc.) that can be manipulated precisely by the experimenter and presented to participants on a computer screen. Such experiments trade away some of the complexity of real-world scenes to allow a maximum of experimental control.


%
We compare five models.
One model is an application (which I will call RSA-REG) of Frank and Goodman's Rational Speech Act (RSA) model \cite{good}, \cite{frank}.\footnote{The mechanisms of \cite{dege} could probably lend RSA-REG better performance \cite{rubi}, but until a systematic performance assessment of the resulting model is available, Frank and Goodman's model will serve our illustrative purposes.} RSA is formalisation of the Gricean idea that communication is always optimally cooperative; consequently, RSA-REG's speaker model emphasizes discriminatory power: the likelihood that a property is chosen for inclusion in a Referring Expression (RE) is proportional to its discriminatory power (i.e., the proportion of scene objects to which the property does {\em not} apply).

The other models grew out of a ``Bounded Rationality" research tradition that emphasises the idea that is skeptical about the idea that speakers routinely compute discriminatory power  when they refer.
A well-known version of this experimentally well-supported idea (e.g. \cite{belk}) is the Incremental Algorithm of \citet{reit}, which assumes that properties are arranged in a linear sequence that lists them according to the degree to which they are preferred. A range of experimental findings \citet{kool, gatt, kees, gomp}) led to various improvements, including two probabilistic versions of the Incremental Algorithm, 
and our own model called Probabilistic Over-specification (PRO), which combines discriminatory power with a probabilistic use of preference.

{\em Comparison 1: Performance.} In  \cite{gomp} we reported an experiment in which the PRO model outperformed the 
other algorithms in terms of the  human-likeness of their output. 

{\em Comparison 2: Direct support.} Algorithms in the Bounded Rationality tradition have often been tested, including the evaluation campaigns of \citet{belz}. Direct support for RSA-REG does not yet reach the level of the other models; naturally, support for these models may grow over time.

{\em Comparison 3: Indirect support.} At first sight, there is much indirect support for RSA, given the intuitive appeal of describing human behaviour as rational. However, a wealth of work in behavioural economics has shown that rational behaviour is affected by time and memory limitations, necessitating shortcuts \cite{elst, simo, gige, gers}, and other deviations from rational behaviour \cite{kahn}. Experiments on REG are in line with these findings \cite{kees,gomp}. For example, PRO is full of shortcuts that avoid the arduous computation of the discriminatory power of each property that would be required by the RSA-REG algorithm. 
%


{\em Comparison 4: Parsimony.} The computational core of RSA-REG can be written in just two simple equations; by contrast, PRO is a rule-based algorithm whose pseudo-code needs about a page. It seems reasonable, therefore, to say RSA-REG is more parsimonious than PRO. 
%

\subsection{Generating Referring Expressions in Context}

{\em REG-in-Context} is another well-studied area of REG. It focusses on co-reference in discourse. It often starts from a text in which all referring expressions (REs) have been blanked out; it predicts, for each of these blanks, what RE should fill it. Other than the identity of the referent, the main information for the model to consider is the sentences around the RE, because this guides the choice between pronouns, proper names, and descriptions. The other entities mentioned in the text play a role not dissimilar to the ``distractor" objects displayed on a computer screen in One-shot REG (previous section).\\[1ex]
A long tradition of linguistic research has led to theories such as accessibility theory \cite{arie}, the givenness hierarchy \cite{gund}, and Centering Theory \cite{bren}. These theories emphasise the effect of the recency of the antecedent (e.g. in terms of the number of intervening words), its animacy (animate/non-animate), and the syntactic structure of the sentences (e.g., Does the RE occur in the same syntactic position as the antecedent?) Computational accounts can be classified in terms of whether they use (1) handwritten rules, (2) hand-coded features and Machine Learning, or (3) an End2End neural architecture.\\[1ex]
Following the GREC evaluation campaign \cite{grec}, in which a number of ML models of REG-in-Context were tested, a wider range of models were recently compared in terms of their performance, looking at human judgments and computational metrics \cite{fafa}. Models included (1) two rule-based ones, RREG-S (small) and RREG-L (small); (2) two models based on traditional Machine Learning (ML), called ML-S (small) and ML-L (large); and (3) three neural models, including two from \citet{cunh} and one from \citet{cao}.

{\em Comparison 1 and 2: Performance and direct support.} Having observed that neural models had only been tested on \citet{ferr}'s version of WebNLG, 
\citet{fafa} decided to test all models 
on WSJ, the Wall Street Journal portion of the OntoNotes corpus \cite{gard}, arguing that WSJ would pose a better test for REG-in-Context algorithms because the texts in it are longer than those in WebNLG.
With respect to WSJ, ML-L outperformed all other models; the simplest rule-based baseline RREG-S performed less well yet it performed at least as well as the neural models on both corpora. 


{\em Comparison 3: Indirect support.} Indirect support varied widely across models, with the larger models receiving the most support from the linguistics literature. RREG-L, for instance, rests on notions such as local focus \cite{bren} and syntactic parallelism \cite{hens}; the large ML model ML-L makes use of the grammatical role of the RE.

The question of Indirect support for neural models is debatable (see below). Unless these models are combined with probing (for the case of REG, see \citet{chen}) or other add-ons, it is difficult to link these models with linguistic insights.\footnote{For the challenge of linking neural models with domain insights, see \citet{kamb}.} On the other hand, neural models may be more inherently cognitively plausible than models based on rules or on classical Machine Learning, because they are inspired by our knowledge of the human brain. Rather than either blithely rejecting or accepting this argument, this is one of those ``higher principles" (section 2) that we should take seriously, while also rigourously investigating their validity (c.f., \cite{ritt}). 


{\em Comparison 4: Parsimony.} Although parsimony can be difficult to assess, some relevant comparisons are straightforward in this case. As observed in \citet{fafa}, the two rule-based models only have the current and previous sentence available to them; the two ML-based models look at the current and all previous sentences; the three neural models have the entire text available to them. The two ``large" models, RREGL and ML-L, contain more rules/features than their smaller counterpart and are consequently less parsimonious. The three models above were architecturally similar {\em seq-2-seq} models with attention in the style of \citet{bahd}, which did not display any obvious differences in terms of parsimony.

\section{Challenges in assessing the explanatory value of a model}

Our case studies illustrate how 
a model may be superior in one respect but inferior in others. And although our first case study suggested a trade-off between parsimony and performance, in which researchers could ``buy" an improvement in performance by sacrificing parsimony, the second case study suggests that this is not always the case. 

On the other hand, some {\em challenges} have come to the fore as well, which I will briefly discuss here. I leave challenges surrounding performance aside here, because they have been widely discussed (e.g., \citet{celi,sai,ehud, spec}) about metrics; \citet{lee} about human evaluation).

{\bf Direct support.} When judging the direct support for a model, younger models tend to be harder to judge, because a younger model cannot be expected to have been subjected to as much scrutiny as an older one, limiting its opportunities for both negative and positive support. 
Where very different results are reported on different corpora (such as the WebNLG and WSJ corpus in \citet{fafa}), further research into the causes of the divergence are in order. 

A further wrinkle in assessing direct support is that models are moving targets: when a model is examined for the second or third time, it is often a modified version of that initial model. What is really being assessed in such cases is not one model but a class of models or, to put it differently, the ideas underlying these models (e.g., that recency and animacy are factors in deciding between the type of RE.

{\bf Indirect support.} Our discussion of rationality put a spotlight on the two ``key questions" that govern indirect support (section 2). For if the relevant principle $I$, as invoked in support of a model, is the idea that behaviour is rational, then some evidence may be available for $I$; but, as it stands, $I$ is too vague to offer strong support for the details of the model, because it does not follow from $I$ that the discriminatory power of the properties in an RE needs to be maximised. Similarly, existing experimental insights do not by themselves dictate all the specifics of the PRO model.

{\bf Parsimony.} 
The idea of parsimony is already well established in NLP practices such as induction of ``causal" models \cite{geig}, knowledge distillation \cite{dist}, and pruning, where the idea is to get rid of parameters or layers that do not add to a model's performance \cite{tess}. Current practices in NLP do not typically involve systematic comparisons between models in terms of their parsimony, and performing such comparisons rigourously is far from trivial. Theoretical equipment for doing so is available in principle, however. Deterministic models, for instance, may be compared with each other in terms of their Kolmogorov complexity \cite{sol2, kolm}; non-deterministic models may be compared in terms of their Minimal Description Length \cite{solo, grun, voit}. Comparisons across different types of models seem more problematic; a complicating factor is that whereas traditional models tend to address one NLP task, neural ``foundation" models such as BERT are adaptable to a wide variety of tasks, which would tend a make a direct comparison across the two types of model biased against foundation models.
%

\section{Policy Implications}

Rather than shying away from them, I believe that our community should embrace the research challenges entailed by an increased emphasis on explanatory value, and the debates that this will bring, including debates about alternative 
dimensions of explanatory value.

Based on my reasoning in the previous sections, I think it would be wrong to limit evaluation of models to only one aspect of their quality. There are parallels here with the assessment of {\em people}, where the influential DORA declaration suggests that the academic community should reduce its reliance on quantitative metrics.\footnote{The San Francisco Declaration On Research Assessment can be found on https://sfdora.org/read/.} Just as academics can have different talents, the success of a model has different dimensions. In both cases, we should learn to juggle multiple dimensions and say things like, {\em ``Based on this experiment, model A has better performance than the older model B. Being relatively new, A still has lower levels of (positive and negative) direct support. However, A is more parsimonious and appears to have better indirect support than B."} 

Policy-wise, researchers and reviewers should be encouraged to think about the explanatory value of models. Analogous to the ethics and limitations paragraphs that are now solicited by some NLP conferences,\footnote{For ethics paragraphs see https://2021.aclweb.org/ethics/Ethics-FAQ/ For limitations paragraphs, https://aclweb.org/portal/content/empirical-methods-natural-language-processing-emnlp-2022.} 
our community could encourage authors of conference papers to comment on all dimensions of explanatory value. Similar moves could be made by institutions that offer funding for scientific research: analogous to letting proposers discuss societal and economic impact, they could be asked to discuss parsimony, and both kinds of support, as well. 
Alternatively, reviewers could be urged to check these dimensions, similar to when reviewers are expected to look out for analyses of statistical significance wherever these are appropriate to the work submitted.

\section{Conclusion}

It is widely accepted that performance alone does not make a good model, because constructing and training models may require a lot of effort; because of concerns over energy consumption; and because of concerns over linguistic, ethnic and other biases (\citet{gebr}). Likewise, novelty and applicability of a model can be important. 

In this Squib, I have argued that another set of dimensions is of crucial importance, particularly when NLP arises from a scientific interest in the world around us; these dimensions, variants of which have often been discussed in connection with other sciences, attempt to make explicit what it means for a model to {\em explain} data. Furthermore, I have argued that explanatory value means more than only performance, and that explanatory value does not equal explainability. 

The borderlines of our discussion are debatable. 
For example, it can be argued that similar arguments apply to NLP-as-Engineering as well. After all, unwieldy models are difficult to maintain and update; models that lack support may fail to generalize, and risk having to be completely redesigned whenever customers' requirements change. Furthermore,
trends in NLP research tend to reflect wider tendencies; accordingly, explanation is taking a backseat in other areas of Artificial Intelligence as well \cite{Leve, kamb}. My conjecture is that the same dimensions of explanatory value, and similar implications for research method and policy, apply there as well. 



\pagebreak

\section{Citations}

\starttwocolumn

\bibliography{squib}

\begin{thebibliography}{72}
\expandafter\ifx\csname natexlab\endcsname\relax\def\natexlab#1{#1}\fi

\bibitem[{Agrawal et~al.(2019)Agrawal, Desai, Wang, Chen, Jain, Johnson, Batra,
  Parikh, Lee, and Anderson}]{agra}
Agrawal, Harsh, Karan Desai, Yufei Wang, Xinlei Chen, Rishabh Jain, Mark
  Johnson, Dhruv Batra, Devi Parikh, Stefan Lee, and Peter Anderson. 2019.
\newblock Nocaps: Novel object captioning at scale.
\newblock In \emph{Proceedings of the IEEE/CVF International Conference on
  Computer Vision}, pages 8948--8957.

\bibitem[{Akmajian and Heny(1975)}]{akma}
Akmajian, Adrian and Frank Heny. 1975.
\newblock \emph{Introduction to the Principles of Transformational Syntax.}
\newblock MIT Press, Cambridge, Mass.

\bibitem[{Alva-Manchego, Scarton, and Specia(2021)}]{spec}
Alva-Manchego, F., C.~Scarton, and L.~Specia. 2021.
\newblock The (un) suitability of automatic evaluation metrics for text
  simplification.
\newblock \emph{Computational Linguistics}, 47(4):861--889.

\bibitem[{Ariel(1990)}]{arie}
Ariel, Mira. 1990.
\newblock \emph{Accessing Noun-Phrase Antecedents}.
\newblock Routledge.

\bibitem[{Bahdanau, Cho, and Bengio(2014)}]{bahd}
Bahdanau, Dzmitry, Kyunghyun Cho, and Yoshua Bengio. 2014.
\newblock Neural machine translation by jointly learning to align and
  translate.

\bibitem[{Belke and Meyer(2002)}]{belk}
Belke, Eva and Antje~S. Meyer. 2002.
\newblock {Tracking the time course of multidimensional stimulus
  discrimination: Analyses of viewing patterns and processing times during
  "same"-"different" decisions}.
\newblock \emph{European Journal of Cognitive Psychology}, 14(2):237--266.

\bibitem[{Belz et~al.(2009)Belz, Kow, Viethen, and Gatt}]{grec}
Belz, Anja, Eric Kow, Jette Viethen, and Albert Gatt. 2009.
\newblock Generating referring expressions in context: The grec task evaluation
  challenges.
\newblock In \emph{Proceedings of ENLG 2009}, pages 294--327, Association for
  Computational Linguistics.

\bibitem[{Bender et~al.(2021)Bender, Gebru, McMillan-Major, and
  Shmitchell}]{gebr}
Bender, Emily~M., Timnit Gebru, Angelina McMillan-Major, and Shmargaret
  Shmitchell. 2021.
\newblock On the dangers of stochastic parrots: Can language models be too big?
\newblock In \emph{Proceedings of the 2021 ACM Conference on Fairness,
  Accountability, and Transparency}, FAccT '21, page 610–623, Association for
  Computing Machinery, New York, NY, USA.

\bibitem[{Berkovitz(2008)}]{berk}
Berkovitz, Joseph. 2008.
\newblock {Action at a Distance in Quantum Mechanics}.
\newblock In Edward~N. Zalta, editor, \emph{The {Stanford} Encyclopedia of
  Philosophy}, {W}inter 2008 edition. Metaphysics Research Lab, Stanford
  University.

\bibitem[{Brennan(1995)}]{bren}
Brennan, Susan~E. 1995.
\newblock Centering attention in discourse.
\newblock \emph{Language and Cognitive processes}, 10(2):137--167.

\bibitem[{Brody(1995)}]{brod}
Brody, Michael. 1995.
\newblock \emph{Lexico-logical form.}
\newblock MIT Press, Cambridge, Mass.

\bibitem[{Cao and Cheung(2019)}]{cao}
Cao, Meng and Jackie Chi~Kit Cheung. 2019.
\newblock Referring expression generation using entity profiles.
\newblock In \emph{Proceedings of the 2019 Conference on Empirical Methods in
  Natural Language Processing and the 9th International Joint Conference on
  Natural Language Processing (EMNLP-IJCNLP)}, pages 3163--3172, Association
  for Computational Linguistics, Hong Kong, China.

\bibitem[{Celikyilmaz, Clark, and Gao(2020)}]{celi}
Celikyilmaz, Asli, Elizabeth Clark, and Jianfeng Gao. 2020.
\newblock Evaluation of text generation: {A} survey.
\newblock \emph{CoRR}, abs/2006.14799.

\bibitem[{Chen, Same, and van Deemter(2021)}]{chen}
Chen, Guanyi, Fahime Same, and Kees van Deemter. 2021.
\newblock What can neural referential form selectors learn?
\newblock In \emph{Proceedings of the 14th International Conference on Natural
  Language Generation}, pages 154--166, Association for Computational
  Linguistics, Aberdeen, Scotland, UK.

\bibitem[{Craver and Tabery(2019)}]{crav}
Craver, Carl and James Tabery. 2019.
\newblock {Mechanisms in Science}.
\newblock In Edward~N. Zalta, editor, \emph{The {Stanford} Encyclopedia of
  Philosophy}, {S}ummer 2019 edition. Metaphysics Research Lab, Stanford
  University.

\bibitem[{Cunha et~al.(2020)Cunha, Ferreira, Pagano, and Alves}]{cunh}
Cunha, Rossana, Thiago Ferreira, Adriana Pagano, and Fabio Alves. 2020.
\newblock Referring to what you know and do not know: Making referring
  expression generation models generalize to unseen entities.
\newblock pages 2261--2272.

\bibitem[{Dale(1989)}]{dale}
Dale, Robert. 1989.
\newblock {Cooking up referring expressions}.
\newblock In \emph{Proceedings of the 27th annual meeting on Association for
  Computational Linguistics (ACL'89)}, pages 68--75, Association for
  Computational Linguistics, Vancouver, BC.

\bibitem[{Dale and Reiter(1995)}]{reit}
Dale, Robert and Ehud Reiter. 1995.
\newblock Computational interpretations of the gricean maxims in the generation
  of referring expressions.
\newblock \emph{Cognitive Science}, 19(2):233--263.

\bibitem[{van Deemter(2016)}]{kees}
van Deemter, Kees. 2016.
\newblock \emph{{Computational Models of Referring: A study in cognitive
  science}}.
\newblock MIT Press, Cambridge, MA.

\bibitem[{Degen et~al.(2020)Degen, Hawkins, Graf, Kreiss, and Goodman}]{dege}
Degen, Judith, Robert~D Hawkins, Caroline Graf, Elisa Kreiss, and Noah~D
  Goodman. 2020.
\newblock When redundancy is useful: A bayesian approach to
  “overinformative” referring expressions.
\newblock \emph{Psychological Review}, 127(4):591.

\bibitem[{Elster(1983)}]{elst}
Elster, Jon. 1983.
\newblock \emph{{Sour Grapes: studies in the subversion of rationality}}.
\newblock MIT Press, Cambridge, MA.

\bibitem[{Ferreira et~al.(2018)Ferreira, Moussallem, Krahmer, and
  Wubben}]{ferr}
Ferreira, Thiago~Castro, Diego Moussallem, Emiel Krahmer, and Sander Wubben.
  2018.
\newblock Enriching the webnlg corpus.
\newblock In \emph{Proceedings of the 11th International Conference on Natural
  Language Generation}, page 171–176, Association for Computational
  Linguistics, Tilburg University.

\bibitem[{Fitzpatrick(2022)}]{fitz}
Fitzpatrick, Simon. 2022.
\newblock Simplicity in the philosophy of science.
\newblock In \emph{Internet Encyclopaedia of Philosophy, ISSN 2161-0002}.

\bibitem[{Frank and Goodman(2016)}]{good}
Frank, Michael~C. and Noah Goodman. 2016.
\newblock {Pragmatic Language Interpretation as Probabilistic Inference}.
\newblock \emph{Trends in Cognitive Sciences}, 20(11):818--829.

\bibitem[{Frank and Goodman(2012)}]{frank}
Frank, Michael~C and Noah~D Goodman. 2012.
\newblock {Predicting pragmatic reasoning in language games.}
\newblock \emph{Science (New York, N.Y.)}, 336(6084):998.

\bibitem[{Gardent et~al.(2017)Gardent, Shimorina, Narayan, and
  Perez-Beltrachini}]{gard}
Gardent, Claire, Anastasia Shimorina, Shashi Narayan, and Laura
  Perez-Beltrachini. 2017.
\newblock Creating training corpora for nlg micro-planners.
\newblock In \emph{Proceedings of the 55th Annual Meeting of the Association
  for Computational Linguistics}, page 179–188, Association for Computational
  Linguistics, Vancouver.

\bibitem[{Gatt et~al.(2013)Gatt, Krahmer, van Gompel, and van Deemter}]{gatt}
Gatt, A, E.~Krahmer, R.P.G. van Gompel, and K.~van Deemter. 2013.
\newblock Factors causing overspecification in definite descriptions.
\newblock In \emph{Proceedings of the 35th Annual Meeting of the Cognitive
  Science Society}.

\bibitem[{Gatt and Belz(2010)}]{belz}
Gatt, Albert and Anya Belz. 2010.
\newblock {Introducing Shared Tasks to NLG: the TUNA Shared Task Evaluation
  Challenges}.
\newblock In Emiel Krahmer and Mariet Theune, editors, \emph{Empirical Methods
  in Natural Language Generation}. Springer.

\bibitem[{Geiger et~al.(2021)Geiger, Lu, Icard, and Potts}]{geig}
Geiger, Atticus, Hanson Lu, Thomas~F Icard, and Christopher Potts. 2021.
\newblock Causal abstractions of neural networks.
\newblock In \emph{Advances in Neural Information Processing Systems}.

\bibitem[{Gershman, Horvitz, and Tenenbaum(2015)}]{gers}
Gershman, S.J., E.J. Horvitz, and J.B. Tenenbaum. 2015.
\newblock Computational rationality: A converging paradigm for intelligence in
  brains, minds, and machines.
\newblock \emph{Science}, 49:273--278.

\bibitem[{Ghassemi, Oakden-Rayner, and Beam(2021)}]{ghas}
Ghassemi, Marzyeh, Luke Oakden-Rayner, and Andrew~L. Beam. 2021.
\newblock The false hope of current approaches to explainable artificial
  intelligence in health care.
\newblock \emph{Lancet Digit Health}, 3:745--750.

\bibitem[{Gigerenzer and Selten(2002)}]{gige}
Gigerenzer, Gerd and Reinhard Selten. 2002.
\newblock \emph{{Bounded Rationality}}.
\newblock MIT Press, Cambridge, MA.

\bibitem[{{V}an Gompel et~al.(2019){V}an Gompel, van Deemter, Gatt, Snoeren,
  and Krahmer}]{gomp}
{V}an Gompel, Roger~PG, Kees van Deemter, Albert Gatt, Rick Snoeren, and
  Emiel~J Krahmer. 2019.
\newblock Conceptualization in reference production: Probabilistic modeling and
  experimental testing.
\newblock \emph{Psychological review}, 126(3):345.

\bibitem[{Greene(2000)}]{gree}
Greene, B. 2000.
\newblock The elegant universe: Superstrings, hidden dimensions, and the quest
  for the ultimate theory.
\newblock \emph{American Journal of Physics}, 68(2):199--200.

\bibitem[{Gruenwald(2007)}]{grun}
Gruenwald, Peter. 2007.
\newblock \emph{The Minimum Description Length Principle}.
\newblock MIT Press.

\bibitem[{Gundel, Hedberg, and Zacharski(1993)}]{gund}
Gundel, Jeanette~K, Nancy Hedberg, and Ron Zacharski. 1993.
\newblock Cognitive status and the form of referring expressions in discourse.
\newblock \emph{Language}, pages 274--307.

\bibitem[{Harnad(1989)}]{harn}
Harnad, Stevan. 1989.
\newblock Minds, machines and searle.
\newblock \emph{Journal of Theoretical and Experimental Artificial
  Intelligence}, (1):5--25.

\bibitem[{Hempel and Oppenheim(1965)}]{hemp}
Hempel, Carl~G. and Paul Oppenheim. 1965.
\newblock Studies in the logic of explanation.
\newblock \emph{Philosophy of Science}, 15(2):135--175.

\bibitem[{Henschel, Cheng, and Poesio(2000)}]{hens}
Henschel, Renate, Hua Cheng, and Massimo Poesio. 2000.
\newblock Pronominalization revisited.
\newblock In \emph{Proceedings of the 18th conference on Computational
  linguistics-Volume 1}, pages 306--312, Association for Computational
  Linguistics.

\bibitem[{Hepburn and Andersen(2021)}]{hepb}
Hepburn, Brian and Hanne Andersen. 2021.
\newblock {Scientific Method}.
\newblock In Edward~N. Zalta, editor, \emph{The {Stanford} Encyclopedia of
  Philosophy}, {S}ummer 2021 edition. Metaphysics Research Lab, Stanford
  University.

\bibitem[{Hodosh, Young, and Hockenmaier(2013)}]{hodo}
Hodosh, Micah, Peter Young, and Julia Hockenmaier. 2013.
\newblock {Framing image description as a ranking task: Data, models and
  evaluation metrics}.
\newblock \emph{Journal of Artificial Intelligence Research}, 47:853--899.

\bibitem[{Hossenfelder(2018)}]{hoss}
Hossenfelder, Sabine. 2018.
\newblock \emph{Lost in Math; How Beauty Leads Physics Astray.}
\newblock Basic Books, New York.

\bibitem[{Hutter, Legg, and Vitanyi(2007)}]{hutt}
Hutter, M., S.~Legg, and P.~M.B. Vitanyi. 2007.
\newblock {A}lgorithmic probability.
\newblock \emph{Scholarpedia}, 2(8):2572.
\newblock Revision \#151509.

\bibitem[{Jaynes(2003)}]{jayn}
Jaynes, E.T. 2003.
\newblock \emph{Probability Theory: The Logic of Science}.
\newblock Cambridge University Press, Cambridge, UK.

\bibitem[{Kahneman and Tversky(2013)}]{kahn}
Kahneman, D. and A.~Tversky. 2013.
\newblock {Prospect theory: An analysis of decision under risk.}
\newblock In \emph{Handbook of the fundamentals of financial decision making.}

\bibitem[{Kambhampati(2021)}]{kamb}
Kambhampati, Subbharao. 2021.
\newblock Polanyi's revenge and ai's new romance with tacit knowledge.
\newblock \emph{Communications of the ACM}, 64(2):31--32.

\bibitem[{Kolmogorov(1965)}]{kolm}
Kolmogorov, A.N. 1965.
\newblock Three approaches to the quantitative definition of information.
\newblock \emph{Problems Inform. Transmission}, 1(1):1--7.

\bibitem[{Koolen et~al.(2011)Koolen, Gatt, Goudbeek, and Krahmer}]{kool}
Koolen, Ruud, Albert Gatt, Goudbeek, and Emiel Krahmer. 2011.
\newblock Factors causing overspecification in definite descriptions.
\newblock \emph{Journal of Pragmatics}, 43(13):3231–3250.

\bibitem[{Krahmer and van Deemter(2012)}]{krah}
Krahmer, Emiel and Kees van Deemter. 2012.
\newblock Computational generation of referring expressions: A survey.
\newblock \emph{Computational Linguistics}, 38(1):173--218.

\bibitem[{van~der Lee et~al.(2019)van~der Lee, Gatt, van Miltenburg, Wubben,
  and Krahmer}]{lee}
van~der Lee, Chris, Albert Gatt, Emiel van Miltenburg, Sander Wubben, and Emiel
  Krahmer. 2019.
\newblock Best practices for the human evaluation of automatically generated
  text.
\newblock In \emph{Proceedings of the 12th International Conference on Natural
  Language Generation}, pages 355--368, Association for Computational
  Linguistics, Tokyo, Japan.

\bibitem[{Lei, Barzilay, and Jaakola(2016)}]{lei}
Lei, Tao, Regina Barzilay, and Tommi Jaakola. 2016.
\newblock Rationalizing neural predictions.
\newblock In \emph{Proceedings of the 2016 Conference on Empirical Methods in
  Natural Language Processing}, pages 107--117.

\bibitem[{Levesque(2014)}]{Leve}
Levesque, H.J. 2014.
\newblock On our best behaviour.
\newblock \emph{Artificial Intelligence}, 212:27--35.

\bibitem[{Lombrozo(2006)}]{lomb}
Lombrozo, Tania. 2006.
\newblock The structure and function of explanations.
\newblock \emph{Trends in Cognitive Sciences}, 10(10):464--470.

\bibitem[{Luo and Shakhnarovich(2017)}]{luo}
Luo, Ruotian and Gregory Shakhnarovich. 2017.
\newblock Comprehension-guided referring expressions.
\newblock In \emph{Proceedings of the IEEE Conference on Computer Vision and
  Pattern Recognition (CVPR)}.

\bibitem[{McKeown(2020)}]{mcke}
McKeown, Kathleen. 2020.
\newblock Rewriting the past: Assessing the field through the lens of language
  generation. (keynote).
\newblock In \emph{Proceedings of the 7th European Workshop on Natural Language
  Generation}.

\bibitem[{Overton(2013)}]{over}
Overton, James~A. 2013.
\newblock “explain” in scientific discourse.
\newblock \emph{Synthese}, (190):1383--1405.

\bibitem[{Popper(1934)}]{popp}
Popper, Karl. 1934.
\newblock \emph{Logik der Forschung, Translated as ``The Logic of Scientific
  Discovery"}.
\newblock Hutchinson, London, 1959.

\bibitem[{Reiter(2018)}]{ehud}
Reiter, Ehud. 2018.
\newblock A structured review of the validity of bleu.
\newblock \emph{Computational Linguistics}, 44(3):393--401.

\bibitem[{Ribeiro et~al.(2020)Ribeiro, Wu, Guestrin, and Singh}]{Ribe}
Ribeiro, Marco~Tulio, Tongshuang Wu, Carlos Guestrin, and Sameer Singh. 2020.
\newblock Beyond accuracy: Behavioral testing of nlp models with checklist.
\newblock In \emph{Proceedings of (ACL 2020)}.

\bibitem[{Ribeiro, Singh, and Guestrin(2016)}]{ribe2}
Ribeiro, M.T., S.~Singh, and C.~Guestrin. 2016.
\newblock Why should i trust you? explaining the predictions of any classifier.
\newblock In \emph{Proceedings of the 22th ACM SSIGKDD Int. Conf. on Knowledge
  Discovery and Data Mining}.

\bibitem[{Ritter et~al.(2017)Ritter, Barrett, Santoro, and Botvinick}]{ritt}
Ritter, Samuel, David G.~T. Barrett, Adam Santoro, and Matt~M. Botvinick. 2017.
\newblock Cognitive psychology for deep neural networks: A shape bias case
  study.

\bibitem[{Rubio-Fernandez(2021)}]{rubi}
Rubio-Fernandez, Paula. 2021.
\newblock Color discriminability makes over-specification efficient:
  Theoretical analysis and empirical evidence.
\newblock \emph{Humanities and Social Sciences Communications}, 8(1):1--15.

\bibitem[{Sai, Mohankumar, and Khapra(2022)}]{sai}
Sai, Ananya~B., Akash~Kumar Mohankumar, and Mitesh~M. Khapra. 2022.
\newblock A survey of evaluation metrics used for nlg systems.
\newblock \emph{ACM Comput. Surv.}, 55(2).

\bibitem[{Same, Chen, and Van~Deemter(2022)}]{fafa}
Same, Fahime, Guanyi Chen, and Kees Van~Deemter. 2022.
\newblock Non-neural models matter: A re-evaluation of neural referring
  expression generation systems.
\newblock In \emph{Proceedings of ACL 2022}, Association for Computational
  Linguistics.

\bibitem[{Sanh et~al.(2019)Sanh, Debut, Chaumond, and Wolf}]{dist}
Sanh, Victor, Lysandre Debut, Julien Chaumond, and Thomas Wolf. 2019.
\newblock Distilbert, a distilled version of {BERT:} smaller, faster, cheaper
  and lighter.
\newblock \emph{CoRR}, abs/1910.01108.

\bibitem[{Simon(1991)}]{simo}
Simon, Herbert. 1991.
\newblock Bounded rationality and organizational learning.
\newblock \emph{Organisational Science}, 2.

\bibitem[{Solomonoff(1964)}]{solo}
Solomonoff, Ray~J. 1964.
\newblock A formal theory of inductive inference: part i.
\newblock \emph{Information and Control}, 7(1):1--22.

\bibitem[{Solomonoff(1960)}]{sol2}
Solomonoff, R.J. 1960.
\newblock \emph{A preliminary report on a general theory of inductive
  inference.}
\newblock Technical Report ZTB-138, Zator Company, Cambridge, Mass.

\bibitem[{Tessier(2021)}]{tess}
Tessier, Hugo. 2021.
\newblock Neural network pruning 101, https://towardsdatascience.com/ \-
  neural-network-pruning-101-af816aaea61.

\bibitem[{Voita and Titov(2020)}]{voit}
Voita, Elena and Ivan Titov. 2020.
\newblock Information-theoretic probing with minimum description length.

\bibitem[{Woodward and Ross(2021)}]{wood}
Woodward, James and Lauren Ross. 2021.
\newblock Scientific explanation.
\newblock In Edward~N. Zalta, editor, \emph{The Stanford Encyclopedia of
  Philosophy}. Springer, pages 264--293.

\bibitem[{Yu et~al.(2016)Yu, Poirson, Yang, Berg, and Berg}]{yu}
Yu, Licheng, Patrick Poirson, Shan Yang, Alexander~C. Berg, and Tamara~L. Berg.
  2016.
\newblock Modeling context in referring expressions.

\end{thebibliography}

\end{document}